\documentclass[10pt,twocolumn,letterpaper]{article}

\usepackage{iccv}
\usepackage{graphicx}
\usepackage{subcaption}
\usepackage{amsmath}
\usepackage{amssymb}
\usepackage{booktabs}
\usepackage{mathtools}
\usepackage{tabularx}

\usepackage{graphicx}
\usepackage{comment}
\usepackage{multirow}
\usepackage{nicefrac}
\usepackage{times}
\usepackage{epsfig}
\usepackage{graphicx}
\usepackage{amsmath}
\usepackage{tabularx}
\usepackage{amssymb}

\usepackage[pagebackref=true,breaklinks=true,letterpaper=true,colorlinks,bookmarks=false]{hyperref}

\iccvfinalcopy % *** Uncomment this line for the final submission

\def\iccvPaperID{66} % *** Enter the ICCV Paper ID here

\ificcvfinal\fi

\begin{document}

%%%%%%%%% TITLE
\title{Deep Learning at the Intersection: Certified Robustness as a Tool for 3D Vision}

\author{Gabriel Pérez S.$^{1*}$, Juan C. Pérez$^{2*}$, \\Motasem Alfarra$^2$, Jesús Zarzar$^2$, Sara Rojas$^2$, Bernard Ghanem$^2$, Pablo Arbeláez$^1$
\and
$^1$Universidad de Los Andes, $^2$KAUST}
\begin{comment}
\author{Gabriel Pérez S\\
Universidad De Los Andes\\
Bogotá, Colombia\\
{\tt\small g.perezsantamaria@uniandes.edu.co}
\and
Juan C. Pérez\\
King Abdullah University of Science and Technology\\
Thuwal, Saudi Arabia\\
{\tt\small author@i2.org}
\and
Motasem Alfarra\\
King Abdullah University of Science and Technology\\
Thuwal, Saudi Arabia\\
{\tt\small author@i2.org}
\and
Jesús Zarzar\\
King Abdullah University of Science and Technology\\
Thuwal, Saudi Arabia\\
{\tt\small author@i2.org}
\and
Sara Rojas\\
King Abdullah University of Science and Technology\\
Thuwal, Saudi Arabia\\
{\tt\small author@i2.org}
\and
Bernard Ghanem\\
King Abdullah University of Science and Technology\\
Thuwal, Saudi Arabia\\
{\tt\small author@i2.org}
\and
Pablo Arbeláez\\
Universidad De Los Andes\\
Bogotá, Colombia\\
{\tt\small pa.arbelaez@uniandes.edu.co}
}
\end{comment}

\maketitle
\def\thefootnote{*}\footnotetext{These authors contributed equally to this work}

\ificcvfinal\thispagestyle{empty}\fi

%%%%%%%%% ABSTRACT
% \begin{abstract}
%    This paper presents an innovative integration of certified robustness principles with 3D scene reconstruction techniques. We propose a method to certify the occupancy predictions of Neural Radiance Fields (NeRF) for calculating weak Signed Distance Functions (SDF) through Randomized Smoothing. By treating the NeRF-derived occupancy grid as a binary classifier, we employ Randomized Smoothing to establish a verifiable weak SDF, offering enhanced reliability in spatial understanding and reconstruction fidelity.
% \end{abstract}

\begin{abstract}
This paper presents preliminary work on a novel connection between certified robustness in machine learning and the modeling of 3D objects. 
We highlight an intriguing link between the Maximal Certified Radius~(MCR) of a classifier representing a space's occupancy and the space's Signed Distance Function~(SDF). 
Leveraging this relationship, we propose to use the certification method of randomized smoothing (RS) to compute SDFs.
Since RS' high computational cost prevents its practical usage as a way to compute SDFs, we propose an algorithm to efficiently run RS in low-dimensional applications, such as 3D space, by expressing RS' fundamental operations as Gaussian smoothing on pre-computed voxel grids. 
Our approach offers an innovative and practical tool to compute SDFs, validated through proof-of-concept experiments in novel view synthesis. 
% The integration of our method with existing novel view synthesis techniques demonstrates promising visual and geometric results. 
This paper bridges two previously disparate areas of machine learning, opening new avenues for further exploration and potential cross-domain advancements.
\end{abstract}
\vspace{-0.55cm}

%%%%%%%%% BODY TEXT
\section{Introduction}
This paper explores preliminary observations on a connection between certified robustness and 3D object modeling in machine learning. 
Certified robustness studies guarantees about model prediction stability under input variations~\cite{raghunathan2018certified, cohen2019certified}, while 3D object modeling focuses on computational representations like meshes, voxel grids~\cite{Tatarchenko_2017_ICCV}, and signed distance functions~(SDFs)~\cite{curless1996volumetric,park2019deepsdf}.

%This paper explores preliminary observations on a connection between two distinct areas in machine learning: certified robustness and the modeling of 3D objects. 
%On the one hand, certified robustness studies theoretical guarantees on the stability of model predictions when inputs are subject to change~\cite{raghunathan2018certified, cohen2019certified}. 
%On the other hand, the modeling of 3D objects focuses on devising useful computational representations, such as meshes, voxel grids \cite{Tatarchenko_2017_ICCV}, and signed distance functions~(SDF)~\cite{curless1996volumetric,park2019deepsdf}.

% Our work centers around a connection we observe between SDFs and a key concept in certified robustness. 
% Specifically, our main theoretical observation is that \textit{computing the Maximal Certified Radius (MCR)}~\cite{Zhai2020MACER:}\textit{ for a classifier \( f \) reflecting spatial occupancy is essentially equivalent to determining the space's SDF.}

\begin{figure}[t]
\includegraphics[width=1\columnwidth]{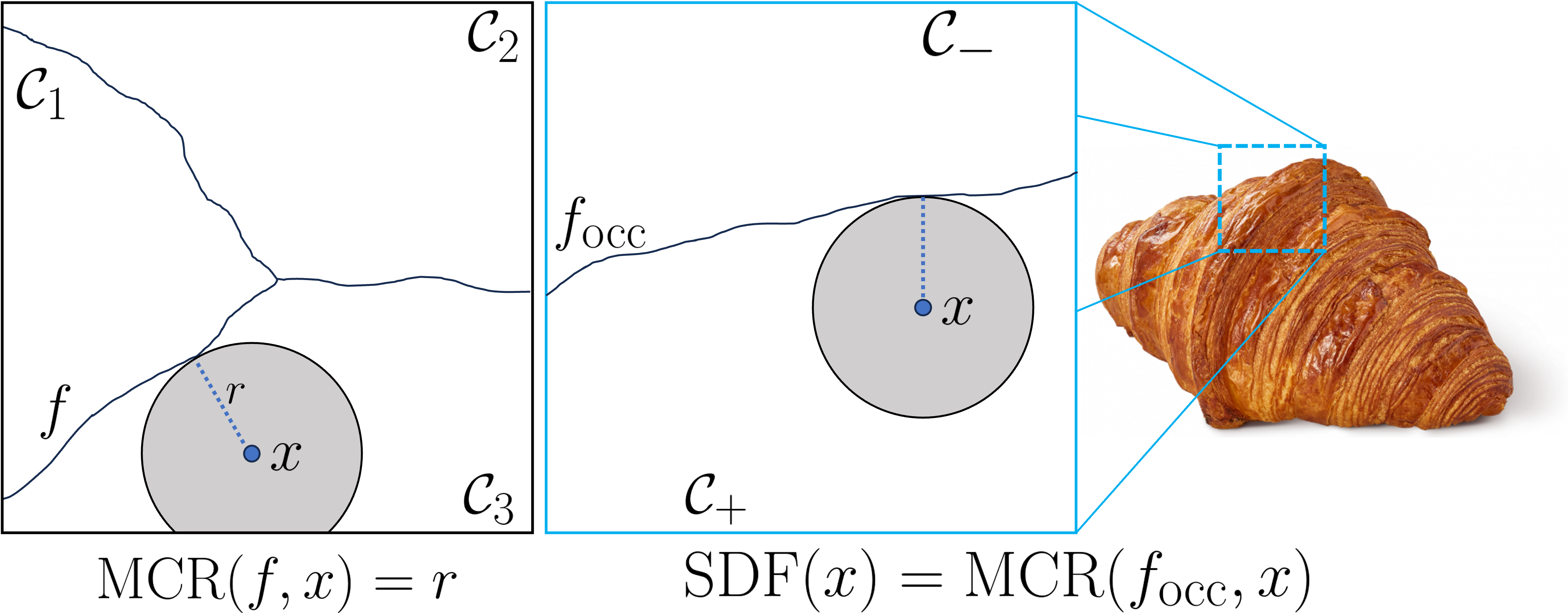}
    \vspace{-0.6cm}
   \caption{\label{fig:pullfig}
   \textbf{Connection between certified robustness and 3D modeling via signed distance function~(SDF).}
   We observe an equivalence between the Maximal Certified Radius~(MCR) of a space's occupancy function $f_\text{occ}$ at a point $x$ and the value of $\text{SDF}(x)$, \textit{i.e.} the (signed) distance to the closest surface.
   }
   \vspace{-0.6cm}
\label{fig:short}
\end{figure}

Our work centers on the link we observe between SDFs and a fundamental concept in certified robustness. 
In particular, we consider the concept of ``Maximal Certified Radius''~(MCR)~\cite{Zhai2020MACER:} for certifying a classifier $f$ at an input $x$: that is, the radius $r$ of the largest ball, around $x$, inside of which $f$'s predictions remain constant.
Based on that concept, our main theoretical observation is that \textit{computing the MCR for a classifier $f$ that represents the space's occupancy is equivalent to computing the space's SDF}.
We visualize this notion in Figure~\ref{fig:pullfig} (left), where we consider computing the MCR of a three-way classifier $f$ at an input $x$, resulting in the radius $r$ for which the prediction of $f$ remains unchanged. 
% Figure ~\ref{fig:pullfig} shows this hypothetical case. 
We note that, if the occupancy in space was modeled via a binary classifier $f_{\text{occ}}$, then the MCR $r$ of $f_{\text{occ}}$ at $x$ corresponds precisely to the distance to the closest surface, which is, by definition, the SDF.

%Please refer to Figure~\ref{fig:pullfig} for a visual guide to our observation.

% Given the connection between SDF and MCR, we explore the certified robustness literature, and find randomized smoothing (RS)~\cite{cohen2019certified} for computing MCRs. 
% However, RS's high computational cost hinders practical SDF computation. To address this, we focus on 3D applications~\cite{Li_2023_CVPR,chabra2020deep,mayost2014applications,park2019deepsdf}, proposing an algorithm that efficiently applies RS in three dimensions through Gaussian smoothing on pre-computed voxel grids.

Given the connection between SDFs and MCRs, we turn to the literature on certified robustness for certification algorithms, where we find the method of randomized smoothing~(RS)~\cite{cohen2019certified}. 
However, we find that RS' high computational cost hinders its usage for computing SDFs in practice. 
To circumvent this, we further make the observation that numerous applications of SDFs are for the 3D world~\cite{Li_2023_CVPR,chabra2020deep,mayost2014applications,park2019deepsdf}, and thus focus our attention on this~(low-dimensional) space.
In this setup, we propose an algorithm that can efficiently run RS in three dimensions by expressing RS' fundamental computations in terms of inexpensive Gaussian smoothing on pre-computed voxel grids.

% \begin{center}
% \end{center}

We validate our preliminary observations via proof-of-concept experiments that test the benefits of using certification to compute SDFs in 3D applications. 
Specifically, we observe that our efficient approach can be integrated into the task of novel view synthesis~\cite{mildenhall2021nerf}, where we model scene geometry through an occupancy grid~\cite{mescheder2019occupancy} whose SDF is easy to compute via our certification algorithm. 
Our experiments show that our method can be added to novel view synthesis techniques to learn a useful representation of the scene while retaining desirable visual results. %  and offers a useful geometric representation of the scene.

Our contributions can be summarized as follows:
\begin{itemize}
    \vspace{-0.2cm}
    \item Identifying a novel link between certified robustness and 3D object modeling, two previously distinct areas in machine learning, providing insights that can benefit both domains.
    \vspace{-0.2cm}
    \item Proposing an algorithm to efficiently compute randomized smoothing certificates, \textit{i.e.} compute SDFs, in low-dimensional applications (such as the physical world) by leveraging Gaussian smoothing on voxel grids.
    \vspace{-0.2cm}
    \item Validating the practical applicability of our approach via proof-of-concept experiments.
    In particular, we find that our algorithm can effectively be used to represent geometry in pipelines for novel view synthesis.
\end{itemize}
\vspace{-0.2cm}
We underscore that this manuscript presents work in progress whose details we are still studying.

\begin{figure*}
% \begin{center}
% \fbox{\rule{0pt}{2in} \rule{.9\linewidth}{0pt}}
\includegraphics[width=1\textwidth]{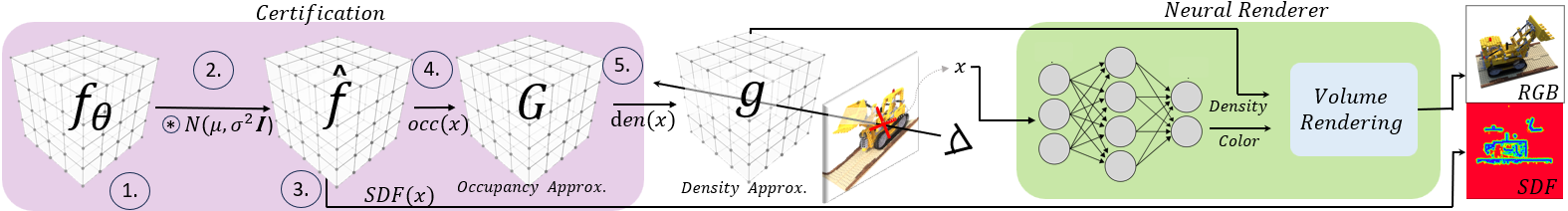}
\vspace{-0.6cm}
% \end{center}
   \caption{\textbf{Rendering pipeline}: A voxel grid \( f_\theta \) is trained within [0, 1]. Gaussian smoothing via 3D convolution produces \( \hat{f} \). Utilizing this, a Weak Signed Distance Function (SDF) incorporates \( \hat{f} \), the normal distribution's CDF, and \( \sigma \). High-eccentricity sigmoid application to \( \hat{f} \) generates \( G(x) \) for occupancy. Rendering density, obtained from \( G(x) \), is calculated using a differentiable density activation function resembling a negative logarithm asymptotically approaching \( 1+\epsilon \), yielding \( g(x) \). The renderer queries \( g(x) \) for density.}

   %\caption{\textbf{Rendering pipeline}: We first train a voxel grid \( f_\theta \) to have values within the 0 to 1 range. Gaussian smoothing is then applied as a 3D convolution to create \( \hat{f} \). From this, a Weak Signed Distance Function (SDF) is obtained by using the transformed \( \hat{f} \), the  CDF of the normal distribution and the smoothing parameter $\sigma$. To achieve hard occupancy classification, a  high-eccentricity sigmoid function is applied to \( \hat{f} \) to generate $G(x)$. Finally, density for rendering is acquired using a differentiable density activation function such as a negative logarithm asymptotic at $1+\epsilon$, which is applied to $G(x)$ to ensure appropriate value alignment thus, producing density field $g(x)$. $g(x)$ may then be queried by the renderer for density at $x$}
\label{fig:short}
\vspace{-0.45cm}
\end{figure*}

\section{Preliminaries: Certified Robustness}
Given the vulnerability of Deep Neural Networks~(DNNs) to small imperceptible perturbations known as adversarial attacks~\cite{goodfellow2015explaining, szegedy2014intriguing}, several works devised approaches with provable robustness guarantees.
That is, given a classifier $f_\theta: \mathcal X \rightarrow \mathbb R^k$ where $\mathcal X$ is the input space and $k$ is the number of classes, certified robustness aims at making $f_\theta$ output a fixed prediction for an $\|\ell\|_p$ region around a given $x$.
In other words, we say that $f_\theta$ is certifiably robust at a given $x$ with radius $r > 0$ if the following statement is achievable:
\vspace{-0.15cm}
\[
\text{arg}\max_if_\theta^i(x) = \text{arg}\max_if_\theta^i(x + \delta) \,\,\,\, \forall \|\delta\| \leq r,
\]
where $f_\theta^i(x)$ is the $i^{\text{th}}$ element of the vector $f_\theta(x)$.
We note that quantifying the certified radius $r$ is, in general, a very challenging problem for DNNs.
However, recently, a tractable approach, known as Randomized Smoothing~\cite{cohen2019certified}, alleviated this problem via a probabilistic approach.

\subsection{Randomized Smoothing}

Randomized Smoothing (RS)~\cite{cohen2019certified} is a technique that constructs a smoothed classifier \(g\) derived from a base classifier $f_\theta$. 
At its core, \(g\) assigns an input \(x\) the class most likely predicted by the base classifier \(f\) when \(x\) is subjected to perturbations with isotropic Gaussian noise. 
More formally, and for a binary classifier $f_\theta : \mathcal X \rightarrow \{0, 1\}$\footnote{Extension to a larger number of classes follows directly (see \cite{cohen2019certified}).}, the smoothed classifier is given by:
\vspace{-0.15cm}
\[ g(x) = \arg\max_{c \in \{0, 1\}} \mathbb{P}_{\epsilon\sim \mathcal N(0, \sigma^2\text{I})}(f_\theta(x + \epsilon) = c).\]
% where \( \delta \sim \mathcal{N}(0, \sigma^2I) \).
where $\sigma^2$ controls the trade-off between robustness and accuracy for the smooth classifier. Cohen \etal~\cite{cohen2019certified} showed that $g$ is certifiably robust at least with radius $R$ given by 
\begin{equation}\label{eq:certified_radius}
    R = \sigma\:\Phi^{-1}(p_A)
\end{equation}
with $\Phi^{-1}$ being the inverse CDF of the standard Gaussian distribution and $p_A = \max_{c \in \{0, 1\}} \mathbb{P}_{\epsilon\sim \mathcal N(0, \sigma^2\text{I})}(f_\theta(x + \epsilon) = c)$.
That is, $g(x) = g(x+\delta) \,\, \forall \|\delta\|_2\leq R$.
While estimating $p_A$ is generally challenging for complex DNNs, Cohen~\textit{et al.} proposed a tractable approach based on Monte-Carlo sampling with confidence bounds~\cite{cohen2019certified}.
Notably, the certified radius provided by RS is exact in the convex case, and a lower bound otherwise.

% The parameter \(\sigma^2\) serves as a hyperparameter for \(g\), mediating a tradeoff between robustness and accuracy. To put it another way, \(g(x)\) returns the class \(c\) for which the decision boundary \({x' \in \mathbb{R}^d : f(x') = c}\) possesses the greatest measure under the distribution \(\mathcal{N}(x, \sigma^2I)\).

% A significant contribution by Cohen et al. provides a tight robustness guarantee for the smoothed classifier \(g\). Based on the Neyman-Pearson lemma, they assert the following: when the base classifier \(f\) classifies samples from \(\mathcal{N}(x, \sigma^2I)\), let the class \(c_A\) be predicted with probability \(p_A = \mathbb{P}(f(x + \delta) = c_A)\), and the next likely class \(c_B\) be returned with probability \(p_B\). The robustness of \(g\) around \(x\) can be characterized by the radius:
% \[ R = \frac{\sigma (\Phi^{-1}(p_A) - \Phi^{-1}(p_B))}{2} \]
% with \(\Phi^{-1}\) being the inverse of the standard Gaussian CDF.

% \textcolor{red}{mention simplification because $k=2$}

% However, direct computation of \(p_A\) and \(p_B\) remains challenging, especially for complex models like deep neural networks. Monte Carlo sampling offers a practical approach, estimating probabilities \(p_A\) and \(p_B\) such that \(p_A \leq \widetilde{p_A}\) and \(p_B \geq \widetilde{p_B}\) with high confidence. The robustness metric still remains valid when substituting \(p_A\) and \(p_B\) with these approximations.

\section{Efficient Randomized Smoothing in Low-dimensional Spaces}\label{sec:efficient_RS}
Let $k \in [0, 1]^d$ be a grid representing the soft occupancy of an object, where $k^i \in [0, 1]$ represents the probability of the $i^{\text{th}}$ voxel being occupied with an object.
Then, $f_\theta(x)$ may continuously represent the  soft occupancy of an object by trilinearly interpolating voxel values of $k$ around $x$.
We observe that the certified radius of each voxel, the distance for the predicted class to flip, represents the Signed Distance Function~(SDF).
Our main aim is to leverage RS to compute the SDF.
However, conducting the Monte-Carlo approach from~\cite{cohen2019certified} on each voxel results in intractable computation.

To that end, we leverage the equivalency between subjecting the input of the base classifier to isotropic Gaussian noise, and convolving the output with a Gaussian distribution~\cite{salman2019provably}.
That is, one can rewrite the smooth classifier as: $g(x) = \arg\max_c \hat f^c(x)$ with
\vspace{-0.15cm}
\[
\hat f(x) = \left(f_\theta * \mathcal N(0, \sigma^2 \text{I})\right)(x),
\]
where $*$ denotes the convolution operator.

While such computation is impractical for classification problems, we unlock its potential in low-dimensional spaces.
In particular, we note that $\hat f(x)$ can be efficiently approximated via inexpensive Gaussian smoothing on a voxel grid discretizing space, which is tractable for low-dimensional spaces.
This formulation allows efficient calculation of certified radii via RS, offering a promising avenue for applying to new domains. %  and tasks where computational efficiency is paramount.

\section{SDF as Certified Radius: Applications in 3D Vision}
Our key insight is the equivalence between a space's Signed Distance Function (SDF) and the Maximal Certified Radius (MCR) of the space's occupancy function. 
In particular, we note that computing the SDF value at a point in space is equivalent to computing the MCR of the occupancy function at that same point. 
Formally, for any given $x$ in a space whose occupancy is described by the occupancy function $f_\text{occ}$, \textit{i.e.} a binary classifier, the following holds:
\begin{equation}\label{eq:equiv}    
\text{SDF}(x) \equiv \text{MCR}(f_\text{occ}, x),
\end{equation}
where \( \text{SDF}(x) \) denotes the signed distance function at point \( x \) and \( \text{MCR}(f_\text{occ}, x) \) denotes the maximal certified radius for the occupancy function at \( x \).
Please refer to Figure~\ref{fig:pullfig} for a visual guide to our observation.
We note that, while Eq.~\eqref{eq:equiv} holds true, the radius given by Eq.~\ref{eq:certified_radius} is, in the general case, a lower bound, and so in practice we estimate a \textit{weak} SDF, \textit{i.e.} a \textit{lower bound} to the SDF.
\subsection{Novel View Synthesis}
An important application in 3D computer vision which can benefit from the use of SDFs is novel view synthesis.
This task consists on predicting novel views of a scene from a set of posed images of the scene.
NeRFs~\cite{mildenhall2021nerf} successfully tackled this task by learning two fields in 3D: a radiance field $\hat{L}_o(\mathbf{x}, \mathbf{d}; \theta) : \mathbb{R}^3 \times \mathbb{R}^2 \to \mathbb{R}^3 $, representing outgoing radiance in each point $\mathbf{x}$ in direction $\mathbf{d}$, and a density field $\sigma(\mathbf{x}; \theta) : \mathbb{R}^3 \to \mathbb{R} $ which captures the scene's geometry.
Given these functions, one can leverage the volume rendering integral
\vspace{-0.5cm}
\begin{equation}
    \hat C(\mathbf{r}; \theta) = \int\limits_{t_n}^{t_f} T(t) \: \sigma(\mathbf{r}(t)) \: \hat{L}_o(\mathbf{r}(t), \mathbf{d}) \: \mathrm{d}t,
\label{eq:volume_rendering}
\end{equation}
% \begin{equation}
%     \:\text{where}\:\: T(t) = \exp\left(- \int\limits_{t_n}^t \sigma(\mathbf{r}(s)) \: \mathrm{d}s\right).
% \label{eq:volume_rendering2}
% \end{equation}
to compute pixel colors $C(\mathbf{r})$ along rays $\mathbf{r}(t)$ in space. 
The parameterized radiance and density functions can be trained by minimizing a reconstruction loss measuring dissimilarity between rendered ray values and original image pixels. 
Please refer to~\cite{mildenhall2021nerf} for further details.

%By approximating the volume rendering equation \ref{eq:volume_rendering} with a discrete approximation it is possible to render color values $ C(\mathbf{r})$ for rays $\mathbf{r}(t)$ through space.
%The radiance and density functions are optimized via a reconstruction loss comparing the rendered ray values with the original image pixels. Refer to NeRF~\cite{mildenhall2021nerf} for more details on this process.

% Deriving the density field from a SDF leads to desirable properties such as a well-defined surface.
% By using SDFs as a geometric representation, several works have improved the reconstruction of geometry from neural fields.
% We test our method by generating an SDF from a predicted occupancy field
We demonstrate an application of our method by generating a density field from the occupancy we learn, and incorporating this density within a popular framework for novel view synthesis~\cite{mueller2022instant}. Our rendering pipeline works as follows:

\vspace{-0.22cm}
%1. \textbf{Learning a Voxel Grid}: We initiate by training a voxel grid to represent a field, which we denote as \( f \). By design, the values of \( f \) lie between 0 and 1.
 
%2. \textbf{Gaussian Smoothing}: Employing a Gaussian kernel with a standard deviation \( \sigma \), we execute Gaussian smoothing on \( f \) to derive its smooth counterpart, \( f_{\text{smooth}} \). Given the constraints on \( f \), \( f_{\text{smooth}} \) invariably retains values within the 0 to 1 range.

%3. \textbf{Deriving the \textit{weak} SDF} : At any point \( x \) of interest, the Signed Distance Function (SDF) can be determined through the equation:
%\[ \text{SDF}(x) = \sigma \times \Phi^{-1}(f_{\text{smooth}}(x)) \]
%where \( \Phi \) denotes the CDF of the Gaussian distribution.

%4. \textbf{Achieving Hard Classification for Occupancy}: While the direct extraction of a hard version from \( f_{\text{smooth}} \) would ideally involve an argmax operation, back-propagation constraints nudge us towards employing a high-eccentricity sigmoid function: 
%\[ \text{occupancy}(x) = \text{sigmoid}\left(\alpha \cdot \left(f_{\text{smooth}}(x) - \nicefrac{1}{2}\right)\right) \]

%5. \textbf{Mapping to Density}: The actual requirement for rendering is the density, which we deduce using an auxiliary, monotonically increasing, and differentiable function \( h \).

%\[h(x) = -30 \cdot \ln(1 + \epsilon - x)\]

\begin{enumerate}
    \item \textbf{Learning a Voxel Grid}: We train a voxel grid, representing a field, which we denote as \( f_\theta \). 
    By design, the values of \( f_\theta \) are clamped between 0 and 1.
    \vspace{-0.15cm}
    \item \textbf{Gaussian Smoothing}: 
    We perform Gaussian smoothing with a kernel of standard deviation \( \sigma \) to derive \( \hat{f} \). 
    Given the constraints on \( f_\theta \), \( \hat{f} \) invariably retains values within the 0 to 1 range.
    \vspace{-0.15cm}
    \item \textbf{Deriving the \textit{weak} SDF}: At any point \( x \) of interest, a weak Signed Distance Function (SDF) can be computed through the equation:
    \vspace{-0.15cm}
    \[ \text{SDF}(x) = \sigma \times \Phi^{-1}\hat{f}(x)) \]
    where \( \Phi \) denotes the CDF of the Gaussian distribution.
    \vspace{-0.5cm}
    \item \textbf{Achieving Hard Classification for Occupancy}: Direct extraction of a hard version from \( \hat{f} \) would involve the argmax operation.
    However, to allow for back-propagation, we approximate the argmax operator with a high-eccentricity sigmoid function
    \[ G(x) = \text{occupancy}(x) = \text{sigmoid}\left(\alpha \cdot \left(\hat{f}(x) - \nicefrac{1}{2}\right)\right). \]
    \vspace{-0.65cm}
    \item \textbf{Mapping to Density}: Rendering requires density, which we model as a (monotonically increasing) transformation of occupancy via a differentiable function \( h \):
    \vspace{-0.25cm}
    \[ g(x) =\text{density}(x) = -30 \cdot \ln(1 + \epsilon - G(x)) \]
\end{enumerate}
\vspace{-0.45cm}

% \paragraph{Maximal certified radius}

% \subsection{Instant-NGP as a Proxy for Grid Initialization}

% To optimize our model's training efficacy, we integrate a "warm-up" stage.

% Starting with the rapid training capabilities of NeRF—thanks in part to the recent advancements like NVIDIA's Neural Graphics Primitives (NGP)—we translate its learned density field to a voxel grid. To mitigate the performance loss by this discretization of space, this grid is finetuned using thir own pipeline using trilinear interpolation to allow continuous queries in space.

% This voxel representation is then inverted through our pipeline to offer a preliminary estimation of the \( f \) grid. Starting from NeRF's learned density voxel grid, we first invert the density activation function, then the occupancy sigmoid function and finally use a 3D adaptation of Wiener deconvolution\cite{} to pseudo invert the gaussian smoothing in space.

% After this inversion step, a few optimization rounds help fine-tune our estimate, so that $f$ closely mirrors NeRF's learned density. 

% Following this stage, the pipeline can be combined directly with NeRF's pipeline, that is, training with the rendering loss through a series of iterations, using trilinear interpolation for rendering equation queries.

\begin{figure*}
    \centering

    \begin{subfigure}[b]{0.15\textwidth}
        \includegraphics[width=0.9\textwidth]{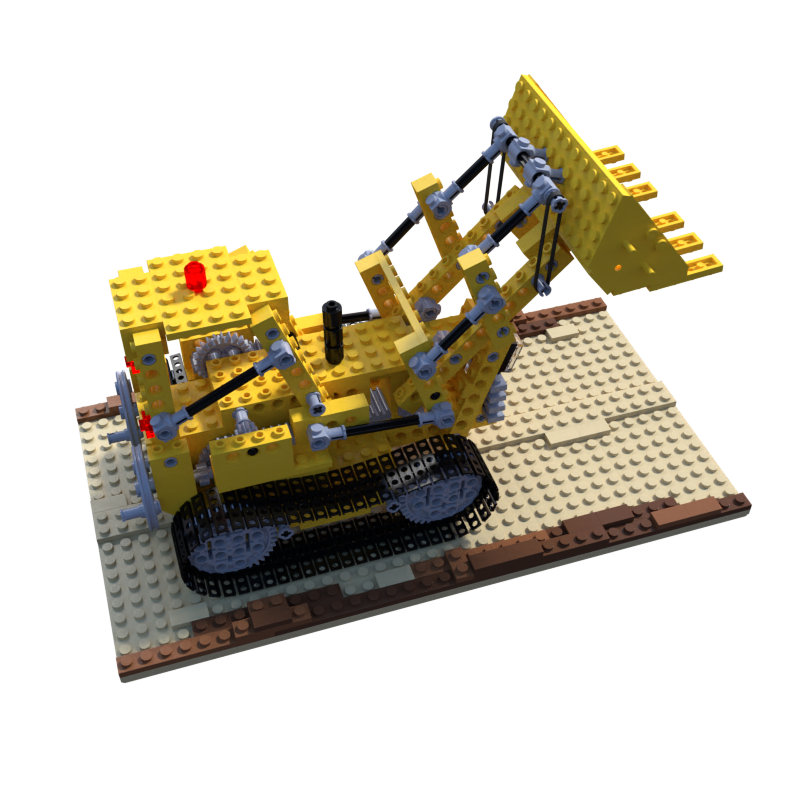}
        \includegraphics[width=0.9\textwidth]{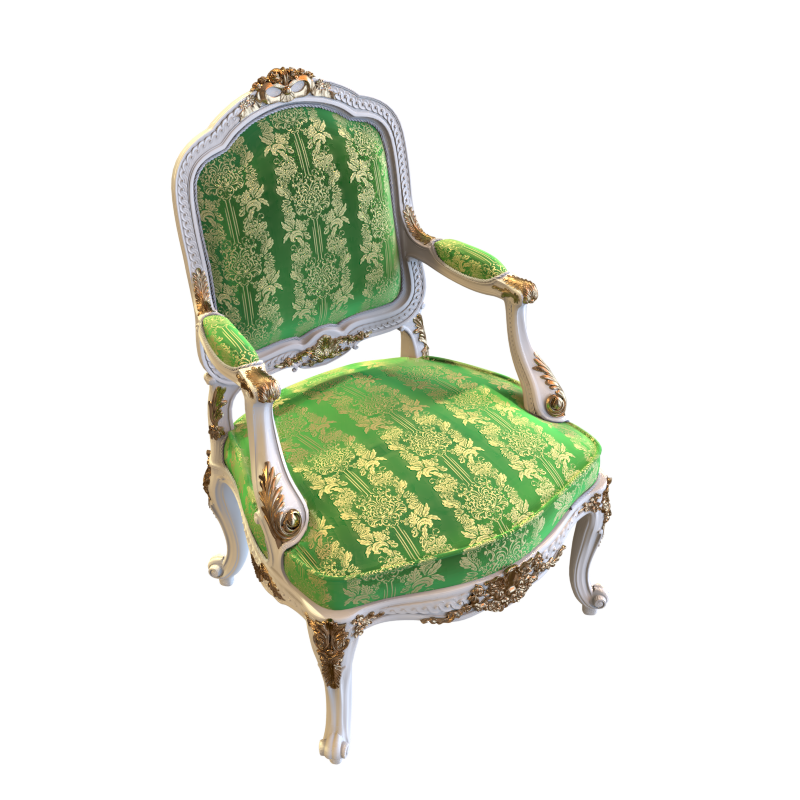}
        \caption{Ground Truth}
    \end{subfigure}%
    \quad
    \begin{subfigure}[b]{0.15\textwidth}
        \includegraphics[width=0.9\textwidth]{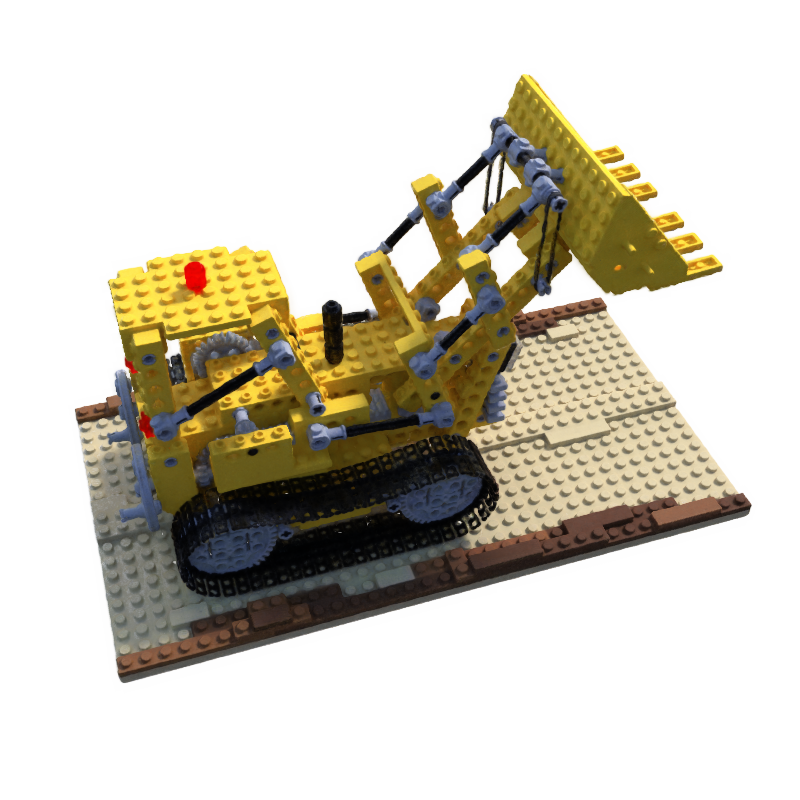}
        \includegraphics[width=0.9\textwidth]{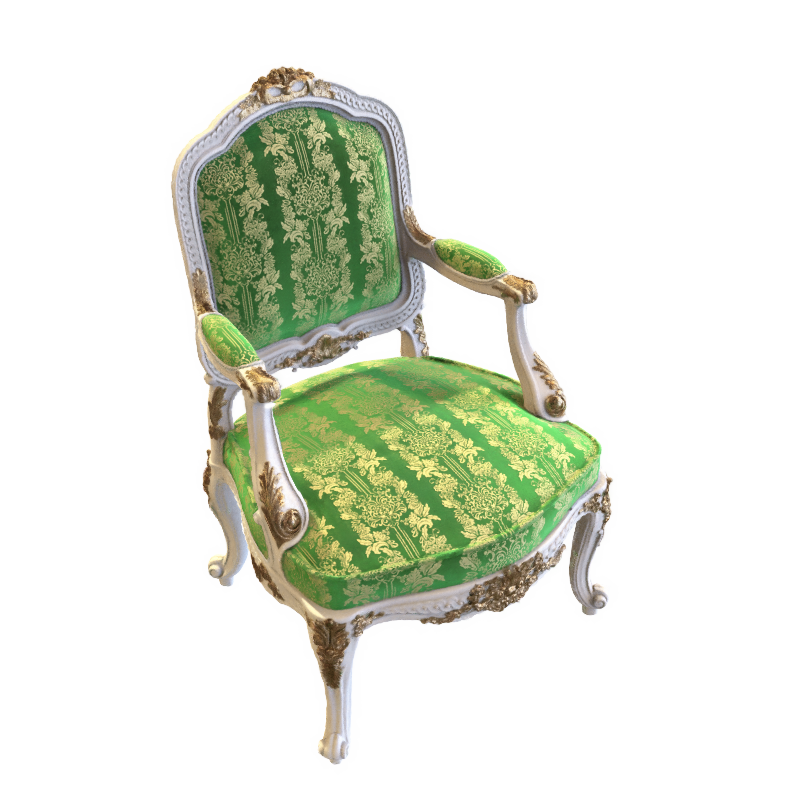}
        \caption{Rendered Image}
    \end{subfigure}%
    \quad
    \begin{subfigure}[b]{0.15\textwidth}
        \includegraphics[width=0.9\textwidth]{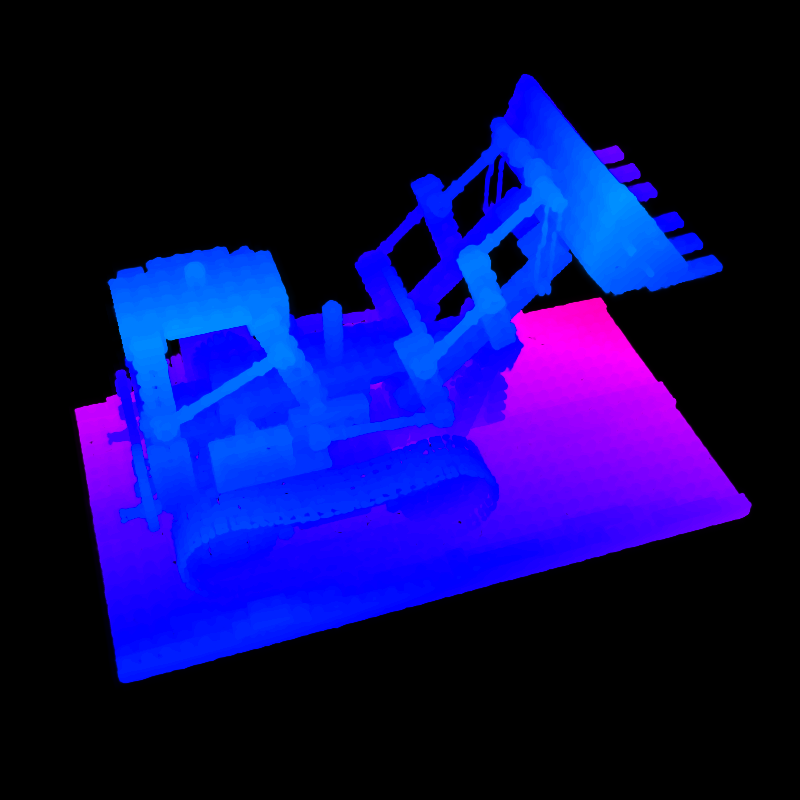}
        \includegraphics[width=0.9\textwidth]{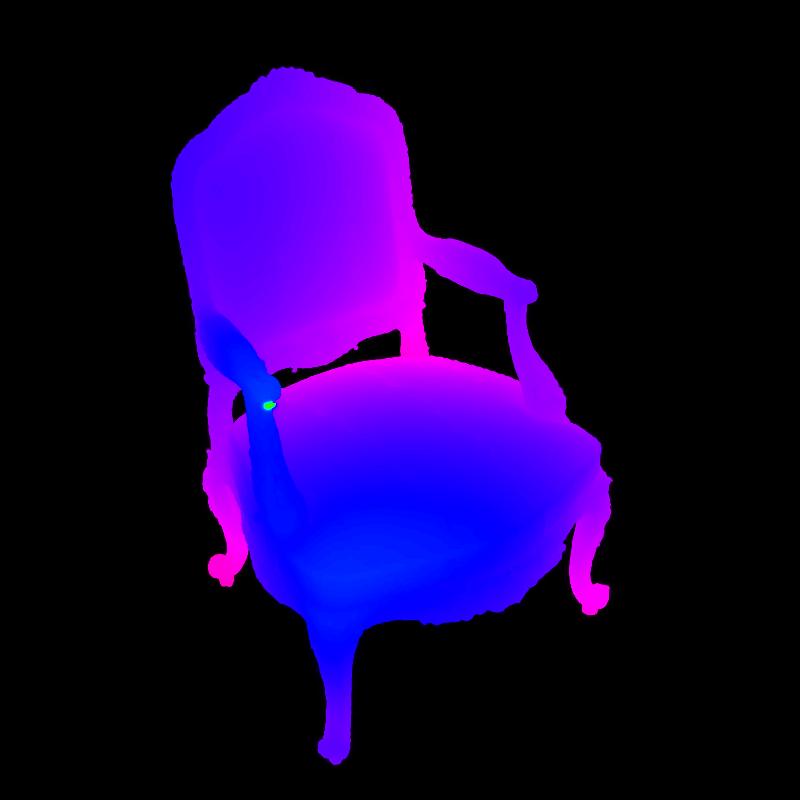}
        \caption{Depth Map}
    \end{subfigure}
    \begin{subfigure}[b]{0.15\textwidth}
        \includegraphics[width=0.9\textwidth]{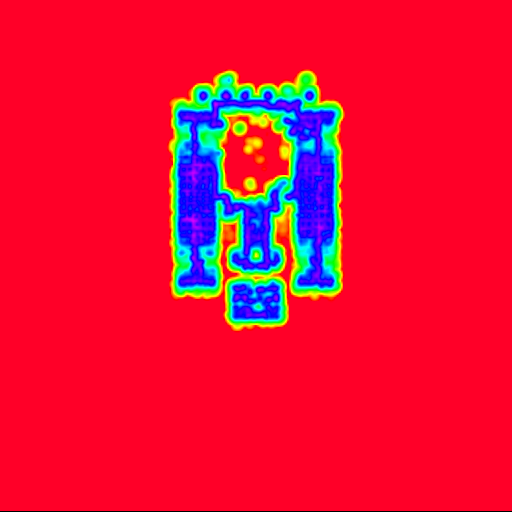}
        \includegraphics[width=0.9\textwidth]{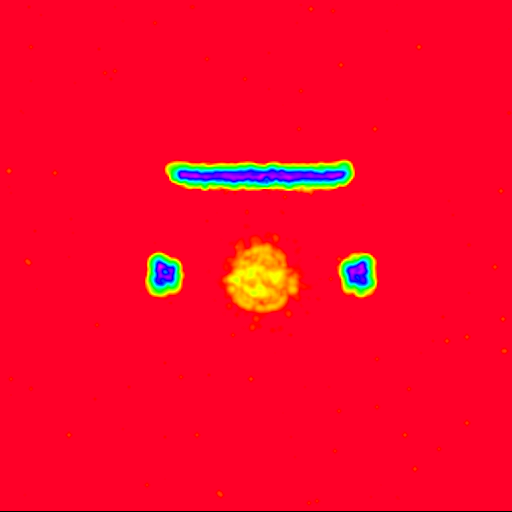}
        \caption{SDF x cut}
    \end{subfigure}
    \begin{subfigure}[b]{0.15\textwidth}
        \includegraphics[width=0.9\textwidth]{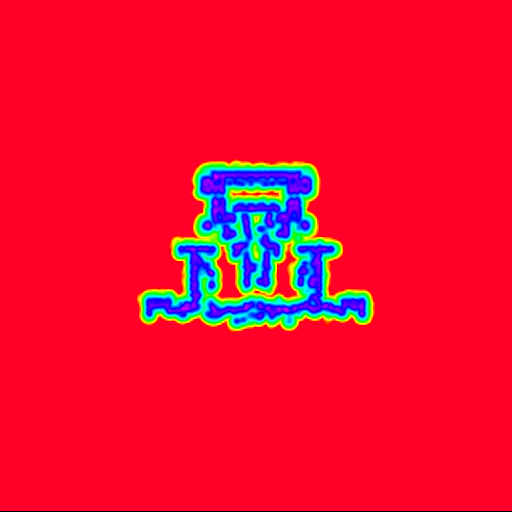}
        \includegraphics[width=0.9\textwidth]{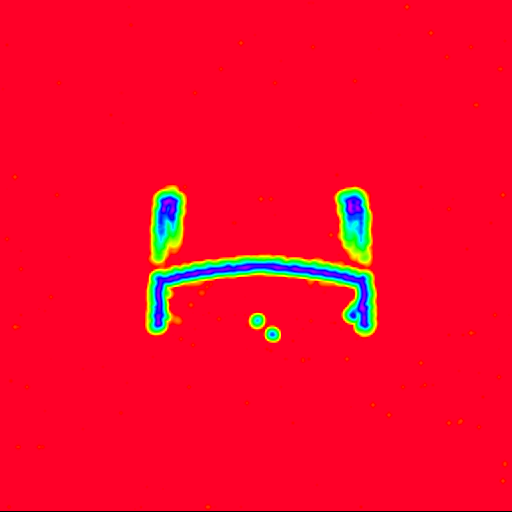}
        \caption{SDF y cut}
    \end{subfigure}
    \begin{subfigure}[b]{0.15\textwidth}
        \includegraphics[width=0.9\textwidth]{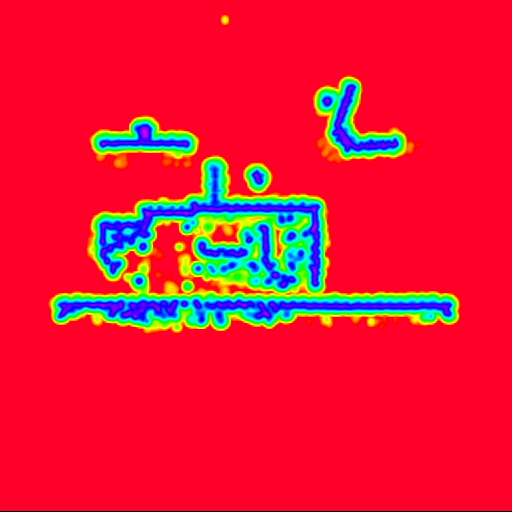}
        \includegraphics[width=0.9\textwidth]{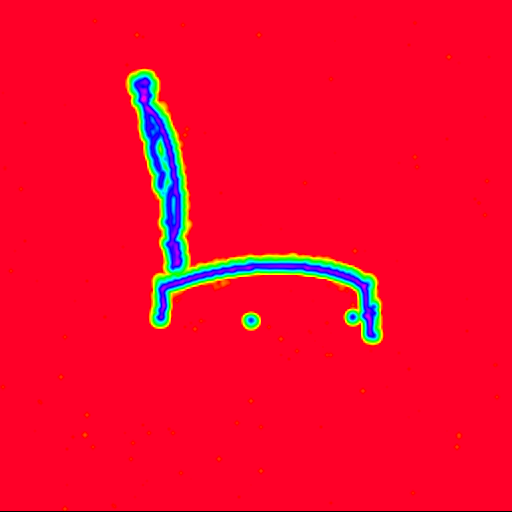}
        \caption{SDF z cut}
    \end{subfigure}
    
    \caption{
    \label{fig:renders,depth,SDFs}
    \textbf{Qualitative results} Ground truth test image, rendered image, radiance field's depth map and SDF axis cuts are presented for the Lego and Chair scenes.
    }
    \vspace{-0.45cm}
\end{figure*}

\section{Experiments \& Results}

\subsection{Setup and Implementation Details}
\vspace{-0.2cm}
\paragraph{Dataset}
We demonstrate our approach using synthetic scenes from NeRF~\cite{mildenhall2021nerf}.
This dataset consists of 8 synthetic scenes with $100$ training images rendered from various poses around the object.
\vspace{-0.45cm}
\paragraph{Implementation} 
We build on top of Nerfacc's~\cite{li2023nerfacc} PyTorch code base~\cite{paszke2019pytorch} implementation of I-NGP~\cite{mueller2022instant} with its default parameters.
% The optimizer, estimator and scheduler are Adam, occupancy grid and multi-step, as used in vanilla Nerfacc\cite{li2023nerfacc}. 
We set the Gaussian smoothing's $\sigma$ to $1.1$, sigmoid's $\alpha$ to 19, use $10$k optimization iterations for I-NGP, $5$k iterations for fine-tuning the voxel grid, and $500$ optimization iterations for the final fine-tuning.
% Discrete grid finetuning learning rate, post inversion finetuning learning rate and full render pipeline training learning rate are set at $9.2e^{-5}$, $6.7e^{-6}$ and $5.8e^{-4}$ for the chair scene; $5.0e^{-4}$, $9.4e^{-7}$ and $2.0e^{-3}$ for the drums scene; $1.4e^{-5}$, $1.5e^{-7}$ and $3.1e^{-3}$ for the ficus scene; $2.4e^{-4}$, $1.1e^{-6}$ and $1.8e^{-3}$ for the hot dog scene; $1.6e^{-5}$, $5.6e^{-6}$ and $2.1e^{-3}$ for the lego scene; $2.8e^{-5}$, $3.4e^{-6}$ and $1.3e^{-3}$ for the materials scene; $4.1e^{-5}$, $5.7e^{-6}$ and $3.1e^{-3}$ for the mic scene and, finally, $2.3e^{-3}$, $2.2e^{-7}$ and $2.3e^{-3}$ for the ship scene.
We test three resolutions for the discrete learned representation \( F \): $192^3$, $256^3$ and $384^3$.% %\(192\times192\times192\), \(256\times256\times256\) and \(384\times384\times384\).
\vspace{-0.45cm}
\paragraph{3D object extraction} 
% Weak SDFs, just like SDFs, contain object surfaces at the 0-level isosurface.
We extract object meshes by employing marching cubes on top of the weak SDF our method generates.
In our experiments, we run marching cubes (MC) searching for a slightly negative isosurface of $\text{SDF}=-0.1$, with the purpose of compensating for locally sharp irregularities in points with certified radii of exactly 0. 
% Then, these meshes can be sampled to obtain a corresponding point cloud of the object.
% The similarity between these point clouds and their corresponding ground truth point cloud evaluated through the Chamfer distance. 
Comparison to I-NGP requires extracting a surface from its density field via MC.
To determine a good density threshold for I-NGP, we run MC with threshold values in multiples of $10$ from $0$ to $100$, and report its best-performing numbers.
\vspace{-0.45cm}
% must be selected as a threshold for density, and points beyond the chosen threshold are deemed solid.
% Marching cubes can then be executed to extract a mesh from the isosurface at the chosen threshold.
% Thresholds of multiples of 10 up to 100 were tested, and the best result is reported.

% In order to process Instant-ngp, one starts with a density field, to get a surface, a threshold for density is selected, that is, an arbitrary density value to use as an anchor to determine whether any queried point in space is empty or solid. After selecting this threshold, marching cubes may be executed to generate a mesh and then sampled just as in our methods case. 
% Thresholds of of multiples of 10 up to 100 were tested and the best result is reported.
\paragraph{Metrics} 
We measure render quality with Peak Noise to Signal Ratio (PSNR) of the generated images, geometry reconstruction via Chamfer distance (with point clouds of $100$k samples), and computational cost via training time. % measure geometry-reconstruction quality, % , and computationa training time.
% The Chamfer distance is measured between point cloud samplings of the ground truth mesh and the mesh extracted from marching cubes.
% Our baseline is the default implementation of Instant-NGP in Nerfacc, and we report the same metrics for this baseline.

\subsection{Results and Analysis}

\begin{table}[]
    \centering
    \resizebox{\linewidth}{!}{
        \begin{tabular}{l|ccccccccc}
            \toprule
            Grid res.  & chair & drums & ficus & hotdog & lego & mat. & mic & ship & avg. \\ \midrule
            192 & 32.65 & 23.75 & 28.96 & 33.83 & 29.74 & 26.84 & 32.50 & 26.66 & 29.36\\
            256 & 32.99 & 24.20 & 30.00 & 34.15 & 30.33 & 26.93 & 32.84 & 27.07 & 29.81\\
            384 & 33.30& 24.60 & 30.89 & 34.08 & 31.01 & 26.60 & 33.10 & 27.17 & 30.09\\
            \bottomrule
            \toprule
            I-NGP  & 35.43 & 25.40 & 33.37 & 36.86 & 35.29 & 29.20 & 36.36 & 29.35 & 32.65\\
            \bottomrule
        \end{tabular}
    }
    \vspace{-0.3cm}
    \caption{
        \label{tab:TestPSNR}
        \textbf{Test set average PSNR ($\uparrow$) on the Blender synthetic dataset.}Our method provides inexpensive access to a \textit{guaranteed} weak SDF while simultaneously providing competitive rendering quality.
    }
    
    \vspace{-0.2cm}
\end{table}

\begin{table}[]
    \centering
    \resizebox{\linewidth}{!}{
        \begin{tabular}{l|ccccccccc}
            \toprule
            Grid res.  & chair & drums & ficus & hotdog & lego & mat. & mic & ship & avg. \\ \midrule
            192 & 661 & 583 & 594 & 614 & 609 & 591 & 678 & 597 & 616\\
            256 & 1216 & 1070 & 1084 & 1104 & 1123 & 1125 & 1262 & 1091 & 1134\\
            384 & 3519& 3116& 3144& 3270& 3131& 3281& 3930& 3173 & 3321\\
            \bottomrule
            \toprule
            I-NGP  & 160 & 148 & 120 & 166 & 153 & 163 & 142 & 156 & 151\\
            \bottomrule
        \end{tabular}
    }
    \vspace{-0.3cm}
    \caption{\label{tab:TrainingTime}
        \textbf{Training time (in seconds).} 
        All experiments were run on a Quadro RTX 8000 GPU.}
    \vspace{-0.2cm}
\end{table}

\begin{table}[]
    \centering
    \resizebox{\linewidth}{!}{
        \begin{tabular}{l|ccccccccc}
            \toprule
            % Grid res.  & chair & drums & ficus & hotdog & lego & mat. & mic & ship & avg.\\ \midrule
            % 192 & 51.70 & 1409.00 & 590.00 & 229.85 & 39.41 & 82.63 & 61.44 & 1639.84 & 512.98\\
            % 256 & 39.68 & 1202.00 &468.66 & 208.50 & 36.01 & 66.40 & 30.88 & 1563.55 & 451.96\\
            % 384 & 39.21 & 754.00 & 87.08 & 110.49 & 33.40  & 45.24& 53.51 & 1607.11 & 341.26\\
            % \bottomrule
            % \toprule
            % I-NGP  & 260.70 & 3954.45 & 1526.43 & 1284.49 & 85.08 & 213.31 & 5974.98 & 1422.54 & 1840.24 \\
            % \bottomrule
            Grid res.  & chair & drums & ficus & hotdog & lego & mat. & mic & ship & avg.\\ \midrule
            192 & 51.7 & 1409.0 & 590.0 & 229.9 & 39.4 & 82.6 & 61.4 & 1639.8 & 513.0\\
            256 & 39.7 & 1202.0 & 468.7 & 208.5 & 36.0 & 66.4 & 30.9 & 1563.6 & 452.0\\
            384 & 39.2 & 754.0 & 87.1 & 110.5 & 33.4  & 45.2 & 53.5 & 1607.1 & 341.3\\
            \bottomrule
            \toprule
            I-NGP  & 260.7 & 3954.5 & 1526.4 & 1284.5 & 85.1 & 213.3 & 5975.0 & 1422.5 & 1840.2 \\
            \bottomrule
        \end{tabular}
    }
    \vspace{-0.3cm}
    \caption{\label{tab:ChamferDistance}
        \textbf{Chamfer distance ($\downarrow$).} 
        % Different density thresholds are used for instant-ngp in each scene, chair:$100$, drums:$100$, ficus:$80$, hotdog:$100$, lego:$100$, materials:$20$, mic:$20$, ship:$100$
        }
    \vspace{-0.45cm}
\end{table}

% \begin{table*}[]
%     \centering
%     \resizebox{\linewidth}{!}{
%         \begin{tabular}{l|ccccccccc}
%             \toprule
%             Grid res.  & chair & drums & ficus & hotdog & lego & mat & mic & ship & avg. \\ \midrule
%             192 & 32.65 / 51.70 / 661 & 23.75 / 1409.00 / 583 & 28.96 / 590.00 / 594 & 33.83 / 229.85 / 614 & 29.74 / 39.41 / 609 & 26.84 / 82.63 / 591 & 32.50 / 61.44 / 678 & 26.66 / 1639.84 / 597 & 29.36 / 512.98 / 616 \\
%             256 & 32.99 / 39.68 / 1216 & 24.20 / 1202.00 / 1070 & 30.00 / 468.66 / 1084 & 34.15 / 208.50 / 1104 & 30.33 / 36.01 / 1123 & 26.93 / 66.40 / 1125 & 32.84 / 30.88 / 1262 & 27.07 / 1563.55 / 1091 & 29.81 / 451.96 / 1134\\
%             384 & 33.30 / 39.21 / 3519 & 24.60 / 754.00 / 3116 & 30.89 / 87.08 / 3144 & 34.08 / 110.49 / 3270 & 31.01 / 33.40 / 3131 & 26.60 / 45.24 / 3281 & 33.10 / 53.51 / 3930 & 27.17 / 1607.11 / 3173 & 30.09 / 341.26 / 3321\\
%             \bottomrule
%             \toprule
%             I-NGP  & 35.43 / 260.70 / 160 & 25.40 / 3954.45 / 148 & 33.37 / 1526.43 / 120 & 36.86 / 1284.49 / 166 & 35.29 / 85.08 / 153 & 29.20 / 213.31 / 163 & 36.36 / 5974.98 / 142 & 29.35 / 1422.54 / 156 & 32.65 / 1840.24 / 151 \\

%             \bottomrule
%         \end{tabular}
%     }
%     \caption{
%         \label{tab:TestPSNR}
%         \textbf{Test set average PSNR $\uparrow$, Chamfer distance $\downarrow$, and training times (s)}
%     }
% \end{table*}

We report proof-of-concept experiments demonstrating our method's capability to generate guaranteed weak SDFs while being able to generate renders of competitive quality in novel view synthesis.

\textbf{Qualitative Results}
We report qualitative results in Figure~\ref{fig:renders,depth,SDFs}.
% To qualitatively see the results of our experiments, we may look at subfigures (a), (b) and (c) of figure \ref{fig:renders,depth,SDFs}. 
We show comparisons between GT test set images, the rendered image with our pipeline, and the depth map of the learned density field. 
These results show that our method is capable of correctly capturing the scene's geometry as well as generating competitive renders. 
% General geometry of the scene as well as details of geometry, color, brightness and shadow details are able to be captured successfully in semantically varied scenes.
We highlight that our method's most important capacity is providing inexpensive access to a weak SDF of the scene. 
We visualize this in subfigures (d), (e) and (f) of Figure~\ref{fig:renders,depth,SDFs} with color-mapped transversal cuts of the values computed by running RS via our efficient algorithm from Section~\ref{sec:efficient_RS}.
% Red indicates points with large certificates in which empty space will not change to occupied space (outside objects), this color gradually slides to blue which indicates points in which occupied space will not change to empty space (inside objects).

% Through the previously mentioned process to sample point clouds from SDFs and density fields, we may observe results of Chamfer distance in Table \ref{tab:ChamferDistance}. First up, the contrast of the average Chamfer distance of Instant-NGP compared to our method is striking. This clearly shows the difference between methods focused on just performing well in Novel view synthesis regardless of the underlying geometry of the scene and those that are constrained to be geometrically aware of the scene represented. Instant-NGP is not an ideal method for 3D reconstruction; different thresholds show how the density field is able to conclude some small isolated "clouds" that can't be seen due to the background but that, when scene is converted to a point cloud, are far from the object and contribute high distances to all points of the ground truth object, thus, extremely high Chamfer distances are to be expected. 

\textbf{SDF Quality}
We report Chamfer distances in Table~\ref{tab:ChamferDistance}.
With respect to this metric, our method displays clear superiority against I-NGP. %  compared to our method is striking and clearly shows the advantage of having an SDF constraint on the scene.
That is, even after choosing the best-performing density threshold for I-NGP for each scene, the average Chamfer distance is usually at least three times as large as that of our approach.

% \textbf{Rendering Quality} Taking a look at table \ref{tab:TestPSNR} we notice an average PSNR of 30.09 when a 384 resolution for $f$ is used, this is a huge numerical giveaway of the aesthetical potential our method maintains. An average drop of 2.51 points of PSNR compared to Instant-ngp can be noticed, however, it is to be remembered the geometrical constraint that our method has in the way that density in space is computed. Taking a deeper dive, we may also notice a trend of higher grid resolution resulting in higher PSNRs, this is to be expected since a higher resolution grid offers more precise density discretization. Nonetheless, the trend seems to indicate diminishing returns as the resolution keeps scaling, even in some scenes like materials and hotdog no benefit can be seen as the resolution increases.
\textbf{Rendering Quality} 
Table~\ref{tab:TestPSNR} reports an average PSNR of 30.09 when using a resolution of 384, which represents a drop of 2.51 points of PSNR compared to I-NGP.
This drop in rendering quality is to be contrasted with the fact that our method \textit{guarantees} the output of a weak SDF. %  and thus highly improved geometry.

%  This correlation between resolution and performance becomes even more important when we take into account Table \ref{tab:TrainingTime}. Space complexity scales cubically and time complexity seems to follow a similar trend. Specifically, on average, training time started at around 10 minutes when using a 192 grid, it almost doubled to 20 minutes when the grid resolution scaled to 256 and then tripled to slightly less than 1 hour when increased to 384. If a balance between training time and performance is to be explored further, it stands to reason for the grid resolution not to go higher since anyway performance does not seem to be increasing by a reasonable margin compared to training time beyond a resolution of 384.
% Table \ref{tab:TrainingTime} also envisions the training time contribution of the base instance of Instant-NGP that is trained for every scene; on average, less than 2 minutes are employed and, as Table \ref{tab:TestPSNR} shows, incredible rendering quality is achieved.

 \textbf{Efficiency}
 The correlation between resolution and both quality and performance becomes critical, as shown Table~\ref{tab:TrainingTime}.
 Space complexity scales cubically, and time complexity seems to follow a similar trend.
 % Specifically, on average, training time started at around 10 minutes when using a 192 grid, it almost doubled to 20 minutes when the grid resolution scaled to 256 and then tripled to slightly less than 1 hour when increased to 384.
 Our results show that, when considering training time against rendering quality, our method rapidly achieves diminishing returns.
 % If a balance between training time and performance is to be explored further, it stands to reason for the grid resolution not to go higher since anyway performance does not seem to be increasing by a reasonable margin compared to training time beyond a resolution of 384.
% Table \ref{tab:TrainingTime} also envisions the training time contribution of the base instance of Instant-NGP that is trained for every scene; on average, less than 2 minutes are employed and, as Table \ref{tab:TestPSNR} shows, incredible rendering quality is achieved.

% Focusing on our method's reconstruction quality, a positive trend between grid resolution and better performance can be seen again. Distance does decrease significantly as grid resolution increases, which is to be expected in general, however, there are scenes like chair and ship that seem to have already reached a saturation point.

% \section{Challenges and Work in Progress}

% \paragraph{Existing approaches try to mimic SDFs}
\vspace{-0.4cm}
\section{Conclusions}

In summary, our work demonstrates a novel connection between certified robustness and 3D object modeling, leading to an efficient algorithm for computing weak SDFs.
% This approach enhances the accuracy and consistency of geometric representation, as showcased for the novel view synthesis task.
We showcase the utility of this connection by employing it for the task of novel view synthesis where it guarantees learning a weak SDF while maintaining rendering quality.
We anticipate that this synergy between robustness and geometry will drive further advancements in machine learning, computer graphics, and related domains, inspiring researchers to explore the untapped potential at this intersection.
% Our trials have revealed that this enhanced procedure accelerates convergence remarkably. More importantly, it matches the rendering quality of the traditional NeRF while also ensuring the learned field is constrained to a valid weak SDF.

% For this citation style, keep multiple citations in numerical (not
% chronological) order, so prefer \cite{Alpher03,Alpher02,Authors14} to
% \cite{Alpher02,Alpher03,Authors14}.

{\small
\bibliographystyle{ieee_fullname}
\bibliography{egbib}

\begin{thebibliography}{10}\itemsep=-1pt

\bibitem{chabra2020deep}
Rohan Chabra, Jan~E Lenssen, Eddy Ilg, Tanner Schmidt, Julian Straub, Steven
  Lovegrove, and Richard Newcombe.
\newblock Deep local shapes: Learning local sdf priors for detailed 3d
  reconstruction.
\newblock In {\em Computer Vision--ECCV 2020: 16th European Conference,
  Glasgow, UK, August 23--28, 2020, Proceedings, Part XXIX 16}, pages 608--625.
  Springer, 2020.

\bibitem{cohen2019certified}
Jeremy Cohen, Elan Rosenfeld, and Zico Kolter.
\newblock Certified adversarial robustness via randomized smoothing.
\newblock In {\em International Conference on Machine Learning (ICML)}, 2019.

\bibitem{curless1996volumetric}
Brian Curless and Marc Levoy.
\newblock A volumetric method for building complex models from range images.
\newblock In {\em Proceedings of the 23rd annual conference on Computer
  graphics and interactive techniques}, pages 303--312, 1996.

\bibitem{goodfellow2015explaining}
Ian Goodfellow, Jonathon Shlens, and Christian Szegedy.
\newblock Explaining and harnessing adversarial examples.
\newblock In {\em International Conference on Learning Representations (ICLR)},
  2015.

\bibitem{Li_2023_CVPR}
Muheng Li, Yueqi Duan, Jie Zhou, and Jiwen Lu.
\newblock Diffusion-sdf: Text-to-shape via voxelized diffusion.
\newblock In {\em Proceedings of the IEEE/CVF Conference on Computer Vision and
  Pattern Recognition (CVPR)}, pages 12642--12651, June 2023.

\bibitem{li2023nerfacc}
Ruilong Li, Hang Gao, Matthew Tancik, and Angjoo Kanazawa.
\newblock Nerfacc: Efficient sampling accelerates nerfs.
\newblock {\em arXiv preprint arXiv:2305.04966}, 2023.

\bibitem{mayost2014applications}
Daniel Mayost.
\newblock {\em Applications of the signed distance function to surface
  geometry}.
\newblock University of Toronto (Canada), 2014.

\bibitem{mescheder2019occupancy}
Lars Mescheder, Michael Oechsle, Michael Niemeyer, Sebastian Nowozin, and
  Andreas Geiger.
\newblock Occupancy networks: Learning 3d reconstruction in function space.
\newblock In {\em Proceedings of the IEEE/CVF conference on computer vision and
  pattern recognition}, pages 4460--4470, 2019.

\bibitem{mildenhall2021nerf}
Ben Mildenhall, Pratul~P Srinivasan, Matthew Tancik, Jonathan~T Barron, Ravi
  Ramamoorthi, and Ren Ng.
\newblock Nerf: Representing scenes as neural radiance fields for view
  synthesis.
\newblock {\em Communications of the ACM}, 65(1):99--106, 2021.

\bibitem{mueller2022instant}
Thomas M\"uller, Alex Evans, Christoph Schied, and Alexander Keller.
\newblock Instant neural graphics primitives with a multiresolution hash
  encoding.
\newblock {\em ACM Trans. Graph.}, 41(4):102:1--102:15, July 2022.

\bibitem{park2019deepsdf}
Jeong~Joon Park, Peter Florence, Julian Straub, Richard Newcombe, and Steven
  Lovegrove.
\newblock Deepsdf: Learning continuous signed distance functions for shape
  representation.
\newblock In {\em Proceedings of the IEEE/CVF conference on computer vision and
  pattern recognition}, pages 165--174, 2019.

\bibitem{paszke2019pytorch}
Adam Paszke, Sam Gross, Francisco Massa, Adam Lerer, James Bradbury, Gregory
  Chanan, Trevor Killeen, Zeming Lin, Natalia Gimelshein, Luca Antiga, Alban
  Desmaison, Andreas Kopf, Edward Yang, Zachary DeVito, Martin Raison, Alykhan
  Tejani, Sasank Chilamkurthy, Benoit Steiner, Lu Fang, Junjie Bai, and Soumith
  Chintala.
\newblock Pytorch: An imperative style, high-performance deep learning library.
\newblock In {\em Advances in Neural Information Processing Systems (NeurIPS)},
  2019.

\bibitem{raghunathan2018certified}
Aditi Raghunathan, Jacob Steinhardt, and Percy Liang.
\newblock Certified defenses against adversarial examples.
\newblock {\em arXiv preprint arXiv:1801.09344}, 2018.

\bibitem{salman2019provably}
Hadi Salman, Jerry Li, Ilya Razenshteyn, Pengchuan Zhang, Huan Zhang, Sebastien
  Bubeck, and Greg Yang.
\newblock Provably robust deep learning via adversarially trained smoothed
  classifiers.
\newblock {\em Advances in Neural Information Processing Systems}, 32, 2019.

\bibitem{szegedy2014intriguing}
Christian Szegedy, Wojciech Zaremba, Ilya Sutskever, Joan Bruna, Dumitru Erhan,
  Ian Goodfellow, and Rob Fergus.
\newblock Intriguing properties of neural networks.
\newblock In {\em International Conference on Learning Representations (ICLR)},
  2014.

\bibitem{Tatarchenko_2017_ICCV}
Maxim Tatarchenko, Alexey Dosovitskiy, and Thomas Brox.
\newblock Octree generating networks: Efficient convolutional architectures for
  high-resolution 3d outputs.
\newblock In {\em Proceedings of the IEEE International Conference on Computer
  Vision (ICCV)}, Oct 2017.

\bibitem{Zhai2020MACER:}
Runtian Zhai, Chen Dan, Di He, Huan Zhang, Boqing Gong, Pradeep Ravikumar,
  Cho-Jui Hsieh, and Liwei Wang.
\newblock Macer: Attack-free and scalable robust training via maximizing
  certified radius.
\newblock In {\em International Conference on Learning Representations}, 2020.

\end{thebibliography}
}

\end{document}